\def\year{2019}\relax
\documentclass[letterpaper]{article} 
\usepackage{aaai19}  
\usepackage{times}  
\usepackage{helvet}  
\usepackage{courier}  
\usepackage{url}  
\usepackage{graphicx}  

\usepackage{subfigure}
\usepackage{booktabs}
\usepackage{amsmath}
\usepackage{color}
\usepackage{algorithm,algorithmicx,algpseudocode}

\usepackage{latexsym}
\usepackage{breakcites}
\usepackage{url}
\usepackage{multirow}
\usepackage{latexsym}
\usepackage{mathrsfs}
\usepackage{amsfonts}
\usepackage{amssymb}
\usepackage{amsmath}
\usepackage{bm}
\usepackage{url}
\usepackage{lipsum}
\usepackage{titlesec}
\usepackage{times}
\usepackage{graphicx} 
\usepackage{subfigure}

\usepackage{natbib}
\usepackage{algorithm,algorithmicx,algpseudocode}
\usepackage{tikz-dependency}
\usepackage{pgfplots}
\usepackage{xcolor}
\usepackage{array}
\usepackage{enumitem}
\usepackage{booktabs}
\usepackage{colortbl}
\usepackage{booktabs}
\usepackage{graphicx} 
\usepackage{arydshln}
\usepackage{multirow}
\usepackage{multicol}
\usepackage{booktabs}

\newcommand{\tabincell}[2]{\begin{tabular}{@{}#1@{}}#2\end{tabular}}
\DeclareMathOperator*{\softmax}{softmax}

\def\W{\mathbf{W}}

\def\h{\mathbf{h}}
\def\rr{\mathbf{r}}

\def\cc{\mathbf{c}}

\def\oo{\mathbf{o}}
\def\u{\mathbf{u}}
\def\g{\mathbf{g}}

\def\ss{\mathbf{s}}
\def\cc{\mathbf{c}}

\def\W{\mathbf{W}}

\def\h{\mathbf{h}}
\def\rr{\mathbf{r}}
\def\x{\mathbf{x}}
\def\cc{\mathbf{c}}

\def\oo{\mathbf{o}}
\def\u{\mathbf{u}}
\def\g{\mathbf{g}}

\newcommand{\task}{\mathcal{T}}

\newcommand{\dataset}{\mathcal{D}}

\newenvironment{itemize*}%
 {\begin{itemize}%
  \setlength{\itemsep}{0pt}%
  \setlength{\parskip}{0pt}}%
 {\end{itemize}}
 \newenvironment{enumerate*}%
 {\begin{enumerate}%
  \setlength{\itemsep}{0pt}%
  \setlength{\parskip}{0pt}}%
 {\end{enumerate}}

\newcommand*\samethanks[1][\value{footnote}]{\footnotemark[#1]}
\begin{document}
%
\title{Learning Multi-task Communication with Message Passing for Sequence Learning}
\title{Multi-task Learning over Graph Structures}

\author{Pengfei Liu$\dag \ddag$, Jie Fu$\ddag$\thanks{These two authors contributed equally}, Yue Dong$\ddag \sharp$\samethanks, Xipeng Qiu$\dag$, Jackie Chi Kit Cheung$\ddag \sharp$\\
 $\dag$School of Computer Science, Fudan University, Shanghai Insitute of Intelligent Electroics \& Systems \\
 \&     $\ddag$MILA  \& $\sharp$McGill University   \\
\{pfliu14,xpqiu\}@fudan.edu.cn, jie.fu@polymtl.ca,yue.dong2@mail.mcgill.ca,jcheung@cs.mcgill.ca}

\maketitle

\begin{abstract}
We present two architectures for multi-task learning with neural sequence models. Our approach allows the relationships between different tasks to be learned dynamically, rather than using an ad-hoc pre-defined structure as in previous work. We adopt the idea from message-passing graph neural networks, and propose a general \textbf{graph multi-task learning} framework in which different tasks can communicate with each other in an effective and interpretable way. We conduct extensive experiments in text classification and sequence labelling to evaluate our approach on multi-task learning and transfer learning. The empirical results show that our models not only outperform competitive baselines, but also learn interpretable and transferable patterns across tasks.
\end{abstract}

\section{Introduction}
Neural multi-task learning models have driven state-of-the-art results to new levels
in a number of language processing tasks, ranging from part-of-speech (POS) tagging \citep{yang2016multi,sogaard2016deep}, parsing \citep{peng2017deep,guo2016exploiting}, text classification \citep{liu2016recurrent,liu2017adversarial} to machine translation \citep{luong2015multi,firat2016multi}.

\begin{figure}[t]
\setlength{\belowcaptionskip}{-0.1cm}
\centering
 \includegraphics[width=0.8\linewidth]{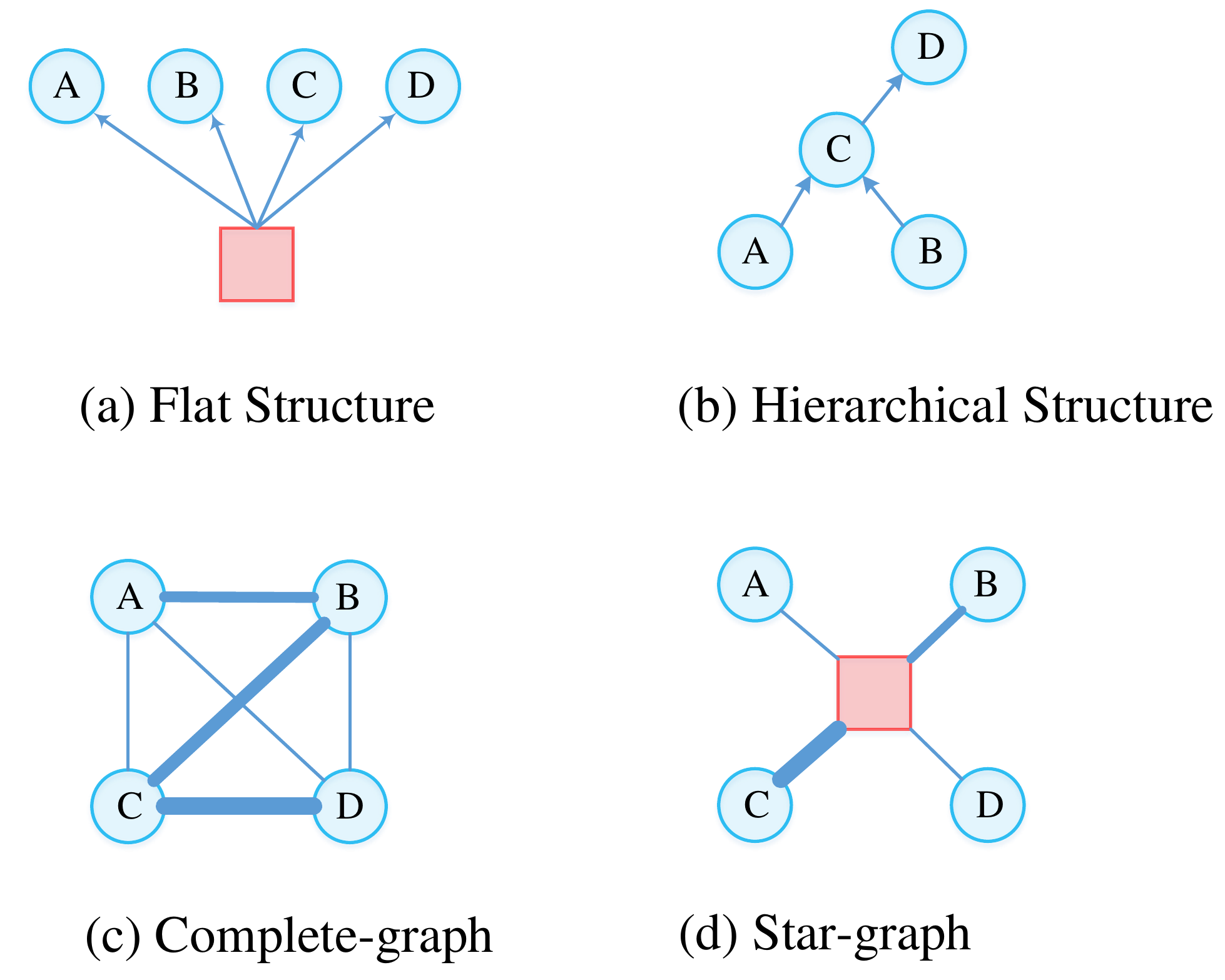}
 \caption{
Different topology structures for multi-task learning: flat, hierarchical and graph structures. Each blue circle represents a task (A, B, C, D), while the red box denotes a virtual node, which stores shared information and facilitates communication. The boldness of the line indicates the strength of the relationships. The directed edges define the direction of information flow. For example, in sub-figure (b), task C receives information from task A, but not vice versa.
}\label{fig:intro}
\end{figure}

Multi-task learning utilizes the correlation between related tasks to improve the performance of each task. In practice, existing work often models task relatedness by simply defining shared common parameters over some pre-defined task structures.
Figure~\ref{fig:intro}-(a,b) shows two typical pre-defined topology structures which have been popular. A \textbf{flat structure} \citep{collobert2008unified} assumes that all tasks jointly share a hidden space, while a \textbf{hierarchical structure} \citep{sogaard2016deep,hashimoto2016joint} specifies a partial order of the direction of information flow between tasks.

There are two major limitations to the above approaches. First, static pre-defined structures represent a strong assumption about the nature of the interaction between tasks, restricting the model's capacity to make use of shared information. For example, the structure in 1(a) does not allow the model to explicitly learn the strength of relatedness between tasks. This restriction prevents the model from fully utilizing and handling the complexity of the data \citep{li2015graph}. Note that the strength of relatedness between tasks is itself not static but subject to change, depending on the data samples at hand. Second, these models are not interpretable to researchers and system developers, meaning that we learn little about what kinds of patterns have been shared besides the parameters themselves.
Previous non-neural-network models \citep{bakker2003task,kim2010tree,chen2010graph} have demonstrated the importance of learning inter-task relationships for multi-task learning. However, there is little work giving an in-depth analysis in the neural setting.

The above issues motivate the following research questions:
1) How can we explicitly model complex relationships between different tasks?
2) Can we design models that learn interpretable shared structures?

To address these questions, we propose to model the relationships between language processing tasks over a \textbf{graph structure}, in which each task is regarded as a node. We take inspiration from the idea of message passing \citep{berendsen1995gromacs, serlet1996method, gilmer2017neural, kipf2016semi}, designing two methods for communication between tasks, in which messages can be passed between any two nodes in a direct (\textit{Complete-graph} in Figure~\ref{fig:intro}-(c)) or an indirect way (\textit{Star-graph} in Figure~\ref{fig:intro}-(d)). Importantly, the strength of the relatedness is learned dynamically, rather than being pre-specified, which allows tasks to selectively share information when needed. 

We evaluate our proposed models on two types of sequence learning tasks, text classification and sequence tagging, both well-studied NLP tasks \citep{li2008multi,liu2017adversarial,yang2016multi}. Moreover, we conduct experiments in both the multi-task setting and in the transfer learning setting to demonstrate that the shared knowledge learned by our models can be useful for new tasks.  Our experimental results not only show the effectiveness of our methods in terms of reduced error rates, but also provide good interpretablility of the shared knowledge.

The contributions of this paper can be summarized as follows:
 \begin{enumerate}
   \item We explore the problem of learning the relationship between multiple tasks and formulate this problem as message passing over a graph neural network.
   \item We present a state-of-the-art approach that allows multiple tasks to communicate dynamically rather than following a pre-defined structure.
   \item Different from traditional black-box learned models, this paper makes a step towards learning \textbf{transferable} and \textbf{interpretable} representations, which enables us to know what types of patterns are shared.
 \end{enumerate}

\section{Message Passing Framework for Multi-task Communication}

We propose to use graph neural networks with message passing to deal with the problem of multi-task sequence learning. Two well-studied sequence learning tasks, text classification and sequence tagging, are used in our experiments. We denote the text sequence as $X = \{x_1,x_2,\ldots, x_T\}$ and the output as $Y$. In text classification, $Y$ is a single label; whereas in sequence labelling,  $Y = \{y_1,y_2,\ldots,y_T\}$ is a sequence.

Assuming that there are $K$ related tasks, we refer to $D_k$ as a dataset with $N_k$ samples for task $k$. Specifically, \begin{equation}
D_k = \{(X_i^{(k)},Y_i^{(k)})\}_{i=1}^{N_k},
\end{equation}
where $X_i^{(k)}$ and $Y_i^{(k)}$ denote a sentence and a corresponding label sequence for task $k$, respectively.
The goal is to learn a neural network to estimate the conditional probability $P(Y | X)$.

Generally, when combining multi-task learning with sequence learning, two kinds of interactions should be modelled: the first is the interactions between different words within a sentence, and the other is the interactions across different tasks. 

For the first type of interaction (\textbf{interaction of words within a sentence}), many models have been proposed by applying a composition function in order to obtain representation of the sentence. Typical choices for defining the composition function include recurrent neural networks \citep{hochreiter1997long}, convolutional neural networks \citep{kalchbrenner2014convolutional}, and tree-structured neural networks \citep{tai2015improved}. In this paper, we adopt the LSTM architecture to learn the dependencies within a sentence, due to their impressive performance on many NLP tasks \citep{cheng2016long}. Formally,
we refer to $\h_t$ as the hidden state of the word at time $t$, $w_t$. Then, $\h_t$ can be computed as:
\begin{align}
\h_t &= \mathbf{LSTM}(\mathbf{x}_t, \h_{t-1}, \theta).  \label{eq:LSTM}
\end{align}
Here, the $\theta$ represents all the parameters of LSTM.

\begin{figure*}[t]
  \centering 
  \subfigure[{Complete-graph MTL}]{
  \includegraphics[width=0.45\linewidth]{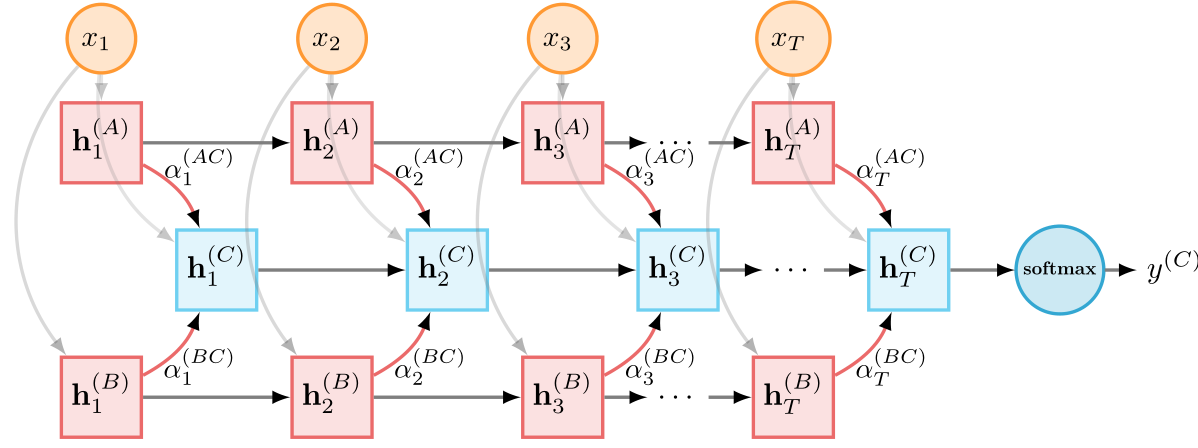}
  }
  \subfigure[Star-graph MTL]{  
  \includegraphics[width=0.4\linewidth]{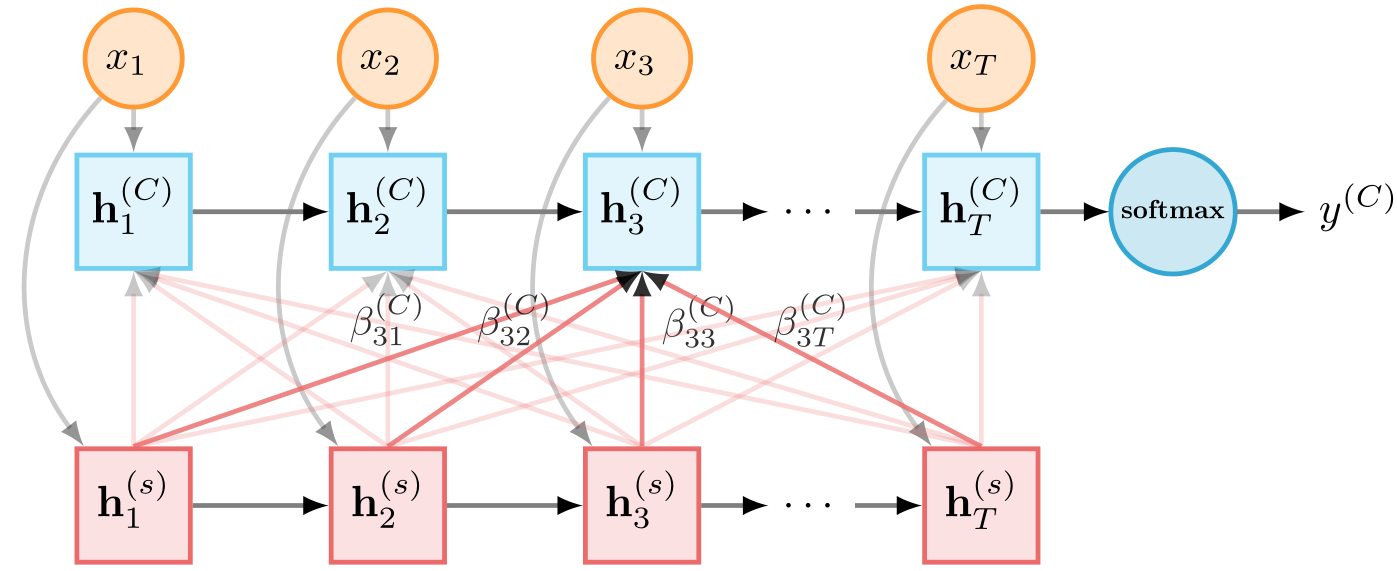}
  }
  \caption{Two frameworks for multi-task communication. (a) shows how task A and B send information to task C. Here, we only present the information flow relating to task C and omit some edges for easy understanding.
  (b) shows how the mailbox (shared layer) sends information to task C.
  The $\alpha$ and $\beta$ values correspond to the edge weights.
  $\alpha$ controls the strength of the association from a word in a source task to a word in the target task while
  $\beta$ controls the amount of information we should take from the shared layer. For example, $\beta_{32}^{C}$  can be understood as: to compute the hidden state $\h_{3}$ of task C, the amount of information should be taken from
  the second hidden state of shared layers is $\beta_{32}^{C}$.
   }\label{fig:framework}
\end{figure*}

For the second type of interaction (\textbf{interaction across different tasks}), we propose to conceptualize tasks and their interactions as a graph, and utilize message passing mechanisms to allow them to communicate. Our framework is inspired by the idea of message passing, which is used ubiquitously in modern computer software \citep{berendsen1995gromacs} and programming languages \citep{serlet1996method}.
The general idea of this framework is that we provide a graph network that allows different tasks to cooperate and interact with one another. Below, we describe our conceptualization of the graph construction process, then we describe the message passing mechanism used for inter-task communication.

\label{sec:graph-construction}
Formally, a graph $G$ can be defined as an ordered pair $G = (V, E)$, where $V$ is a set of nodes $\{V_1,\ldots,V_m\}$ and $E$ is a set of edges. In this work, we use directed graphs to model the communication flows between tasks and an edge is therefore defined as an ordered set of two nodes $(V_i,V_j), i \neq j$.

In our models, we represent each task as a node. In addition, we allow virtual nodes to be introduced. These virtual nodes do not correspond to a task. Rather, their purpose is to facilitate communication among different tasks. Intuitively, the virtual node functions as a mailbox, storing messages from other nodes and distributing them as needed.

Tasks and virtual nodes are connected by weighted edges, which represent communication between different nodes. Previous flat and hierarchical architectures for multi-task learning can be considered as graphs with fixed edge connections. Our models dynamically learn the weight of each edge, which allows the models to adjust the strength of the communication signals.

\subsection{Message Passing}
\label{sec:message-passing}
In our graph structures, we use directed edges to model the communication between tasks. In other words, nodes communicate with each other by sending and receiving messages over edges. Given a sentence with a sequence of words $(w_1^{k}, ..., w_T^{k})$ from task $k$,  we use $\rr_{t}^k$ to represent the \textbf{aggregated messages} that the word of task $k$ at time step $t$ can get, and we use $\h_{t}^{k}$ to denote the task-dependent hidden representation of the word $w_t^k$.

Below, we propose two basic communication architectures for message passing: \textit{Complete-graph} and \textit{Star-graph}  which differ according to whether they allow direct communication between any pair of tasks, or whether the communication is mediated by an intermediate virtual node.

\textbf{1. Complete-graph (CG): Direct Communication for Multi-Task Learning:}
in this model, each node can directly send (or receive) messages to (or from) any other nodes.
Specifically, as shown in Fig.\ref{fig:framework}-(a), we first assign each task a task-dependent LSTM layer. Each sentence in task $k$ can be passed to all the other task-dependent LSTMs\footnote{At training time, the loss is only calculated and used to compute the gradient for the task from which the sentence is drawn.} to get corresponding representations $\h_t^{(i)}$, $i=1 ... K$, $i \neq k$. Then, these messages will be aggregated
as:
\begin{align}
\label{eq:dc_mtl}
    \rr_{t}^{(k)} = \sum_{i=1 ... K \textit{s.t.} i \neq k} \alpha^{(i\rightarrow k)}_{t} \h_t^{(i)}
\end{align}
Here, $\alpha^{(ki)}_{t}$ is a scalar, which controls the relatedness between two tasks $k$ and $i$, and can be dynamically computed as:
\begin{align}
\ss_{t}^{(i\rightarrow k)} &= f(\x_t^{(k)}, \h_{t-1}^{(k)}, \h_{t}^{(i)}) \\
         &= \mathbf{u}^{(s)} \tanh(\W^{(s)}[\x_t,\h_{t-1}^{(k)}, \h_{t}^{(i)}]) \label{eq:dc-attention}
\end{align}
where $\u^{(s)}$ and $\W^{(s)}$ are learnable parameters.
And the relatedness scores will be normalized into a probability distribution:
\begin{align}
 \bm{\alpha}_{t} &= \softmax({\ss_{t}}) \label{eq:dc-softmax}
\end{align}

\textbf{2. Star-graph (SG): Indirect Communciation for Multi-Task Learning:}
the potential limitation of our proposed CG-MTL model lies in its computational cost, because the number of pairwise interactions grows quadratically with the number of tasks.
Inspired by the mailbox idea used in the traditional message passing paradigm \citep{netzer1995optimal}, we introduce an extra virtual node into our graph to address this problem. In this setting, messages are not sent directly from one node to another, but are bridged by the virtual node.
Intuitively, the virtual node stores the shared messages across all the tasks; different tasks can put messages into this global space, then other tasks can take out the useful messages for themselves from the same space.
Fig.\ref{fig:framework}-(b) shows how one task collects information from the mailbox (shared layer).

In details, we introduce an extra LSTM to act as the virtual node, whose parameters are shared across tasks.
Given a sentence from task $k$, its information can be written into the shared LSTM by the following operation:
%
\begin{align}
    \h_t^{(s)}  & =\mathbf{LSTM}(x_t^{(k)}, \h_{t-1}^{(s)}, \theta^{(s)}),
\end{align}
where $\theta^{(s)}$ denotes the parameters are shared across all the tasks.

Then, the aggregated messages at time $t$ can be read from the shared LSTM:
\begin{align}
\label{eq:shared_lstm}
    \rr_{t}^{(k)} = \sum_{i=1}^{T} \beta^{(k)}_{t\rightarrow i} \h^{(s)}_{i},
\end{align}
where $T$ denotes the length of the sentence and $\beta_{t\rightarrow i}^{(k)}$ is used to retrieval useful messages from the shared LSTM, which can be computed in a similar fashion to Equations~\ref{eq:dc-attention} and \ref{eq:dc-softmax}.

Once the graphs and message passing between the nodes are defined, the next question to ask is how to update the task-dependent representation $\h_t^{(k)}$ for node $k$ using the current input information $\x_t$ and the aggregated messages $\rr_{t}^{(k)}$. 
We employ a gating unit that allows the model to decide how many aggregated messages should be used for the target tasks, which avoids unnecessary information redundancy.
Formally, the $\h_t^{(k)}$ can be computed as:
\begin{align}
    \h_t^{(k)}  & =\mathbf{LSTM}^{\dag}(\x_t, \h_{t-1}^{(k)}, \rr_{t}^{(k)}, \theta^{{(k)}}, \theta^{{(s)}}).
\end{align}
The function $\mathbf{LSTM}^{\dag}$ is the same as Eq.\ref{eq:LSTM} except that we replace the memory update step of the inner function in Eq.\ref{eq:LSTM} with the following equation:
%
\begin{align}
\h_t &=\oo_t \odot \tanh(\cc_t + \g_t \odot (\W^{(s)}_f \rr_{t})),
\end{align}
where $\W^{(s)}_f$ is a parameter matrix, and $\g_t$ is a fusion gate that selects the aggregated messages. $\g_t$ is computed as follows:
\begin{align}
\g_t = \sigma(\W^{(s)}_r \rr_{t}^{(k)} + \W^{(s)}_c \mathbf{c}_t),
\end{align}
where $\W^{(s)}_r$ and $\W^{(s)}_c$ are parameter matrices.

 \paragraph{Comparison of Complete-graph (CG) and Star-graph (SG)}
 For CG-MTL,  the advantage is that we can figure out the strength of the association from a word in a source task to a word in the target task. However, the computation of CG-MTL is not efficient if the number of tasks is too large.
 For SG-MTL, the advantage is that the learned shared structures are interpretable and more importantly, the learned knowledge of SG-MTL can be used for unseen tasks.
 To conclude, the CG-MTL can be used in these scenarios:
 1) The number of tasks is not too large;
 2) We need to explicitly analyze the relatedness between different tasks (as shown in Fig.3).
 By contrast, the SG-MTL can be used in the following scenarios:
 1) The number of tasks is  large;
 2) We need to  transfer shared knowledge to new tasks;
 3) We need to analyze what types of shared patterns have been learned by the model (As shown in Tab.2).

\subsection{Task-dependent Layers}
Given a sentence $X$ from task $k$ with its label $Y$ (note $Y$ is either a classification label or sequential labels) and its feature vector $\h^{(k)}$ emitted by the two communication methods above, we can adapt our models to  different tasks by using different task-specific layers. We call the task-specific layer as the \textsc{Output-layer}. For text classification tasks, the commonly used \textsc{Output-layer} is a softmax layer, while for sequence labelling tasks, it can be a  conditional random field (CRF) layer. Finally, the output probability $P(Y|X)$ can be computed as:
\begin{align}
P(Y|X) = \textsc{Output-layer}(X,\h^{(k)}_T,\theta^{(k)})\label{eq:softmax}.
\end{align}

Then, we can maximize the above probability to optimize the parameters of each task:
\begin{align}
\mathcal{L}_{k}(X,Y,\theta^{(k)}, \theta^{(s)} ) =  - P(Y|X).
\end{align}


Once the loss function for a specific task is defined, we could compute the overall loss for our multi-task models as the following:
\begin{align}
\mathcal{L} &= \sum_{k=1}^{K}\lambda_k \mathcal{L}_{k}(X,Y,\theta^{(k)}, \theta^{(s)} )
\end{align}
where $\lambda_k$ is the weight for task $k$.
The parameters of the whole network are trained over all datasets and the overall training procedure is presented in Algorithm~\ref{alg:1}.

\begin{algorithm}[t]
\caption{Training Process for Multi-task Learning over Graph Structures}  \label{alg:1}
\label{alg:metanet}
\begin{algorithmic}[1]
{\footnotesize
\Require A set of training tasks $\lbrace\dataset^{\task_i}\rbrace^K_{i=1}$, where $\task_i $ is drawn from $p(\task)$
\State Initialize $\Theta := \lbrace \theta^{(s)}, \theta^{(k)} \rbrace $

\While{not done}
    \For{$\task_k\thicksim p(\task)$}
        \State{Sample a batch of samples  $\mathcal{B}^{\task_k} = \lbrace X,Y\rbrace \in \dataset^{\task_k}$}

        \State  \textcolor[rgb]{0.00,0.59,0.00}{{// \textit{Message Passing}}}

        \If{\texttt{Direct-Communication}}
            \For{$\task_j\thicksim p(\task)$}  \Comment{\textcolor[rgb]{0.00,0.59,0.00}{$\task_k \neq \task_j$}}
            \State{ $\h_t^{(j\rightarrow k)} = \mathbf{LSTM}(\mathcal{B}^{\mathcal{T}_k}, \h_{t-1}, \theta^{(k)})$}
            \EndFor
            \State{$\rr_{t}^{(k)} = \sum_{j}\alpha^{(j\rightarrow k)}_{t} \h_t^{(j)}$}
            \Comment{\textcolor[rgb]{0.00,0.59,0.00}{Aggregating}}
        \ElsIf{\texttt{Indirect-Communication}}
            \State{$\h_t^{(s)} = \mathbf{LSTM}(\mathcal{B}^{\mathcal{T}_k}, \h_{t-1}^{(s)},\theta^{(s)})$}
            \State{$\rr_{t}^{(k)} = \sum_{i=1}^{T} \beta^{(k)}_{i
            \rightarrow t} \h^{(s)}_{i}$}
            \Comment{\textcolor[rgb]{0.00,0.59,0.00}{Aggregating}}
        \EndIf

        \State  \textcolor[rgb]{0.00,0.59,0.00}{{// \textit{Node Updating}}}
        \State{$\h_t^{(k)}   =\mathbf{LSTM}^{\dag}(\x_t, \h_{t-1}^{(k)}, \rr_{t}^{(k)}, \theta^{{(k)}}, \theta^{{(s)}})$}

        \State  \textcolor[rgb]{0.00,0.59,0.00}{{// \textit{Task-dependent Output}}}
        \State{$P(Y|X) = \textsc{Output-layer}(X,\h^{(k)}_T,\theta^{(k)})$}

    \EndFor
\EndWhile

}
  \end{algorithmic}
\end{algorithm}

\section{Experiments and Results}

In this section, we describe our hyperparameter settings and present the empirical performance of
our proposed models on two types of multi-task learning datasets, first on text classification, then on sequence tagging. Each dataset contains several related tasks.

\subsection{Hyperparameters}
The word embeddings for all of the models are initialized with the 200-dimensional GloVe vectors (840B token version \citep{pennington2014glove}).
The other parameters are initialized by randomly sampling from the uniform distribution of $[-0.1, 0.1]$.
The mini-batch size is set to 8.


For each task, we take the hyperparameters which achieve the best performance on the development set via a grid search over combinations of the hidden size $[100, 200, 300]$ and  $l_2$ regularization $[0.0, 5E-5,1E-5]$.
Additionally, for text classification tasks, we set an equal lambda for each task; while for tagging tasks, we run a grid search of lambda in the range of $[1, 0.8, 0.5]$ and take the hyperparameters which achieve the best performance on the development set.
Based on the validation performance, we choose the size of hidden state as $200$ and $l_2$ as 0.0.
We apply stochastic gradient descent with the diagonal variant of AdaDelta for optimization \citep{zeiler2012adadelta}.

\subsection{Text Classification}
To investigate the effectiveness of multi-task learning, we experimented with 16 different text classification tasks involving different popular review corpora, such as books, apparel and movie \citep{liu2017adversarial}. Each sub-task aims at predicting a correct sentiment label (positive or negative) for a given sentence.
All the datasets in each task are partitioned into  training, validating, and testing with the proportions of 1400, 200 and 400 samples respectively.




\begin{table*}[!t]\small
\center
\begin{tabular}{l*{9}{c}}
\toprule
\multirow{2}{*}{\textbf{Task}}  & \multicolumn{1}{c}{\textbf{Single Task}} & \multicolumn{5}{c}{\textbf{Multiple Tasks}} & \multicolumn{2}{c}{\textbf{Transfer}}\\
\cmidrule(lr){2-2}  \cmidrule(lr){3-7} \cmidrule(lr){8-9}
 &  Avg. & MT-CNN & FS-MTL & SP-MTL & CG-MTL* & SG-MTL* & SP-MTL & SG-MTL* \\
\midrule
\multicolumn{1}{l}{Books}            & 19.2   & 15.5$_{(-3.7)}$ & 17.5$_{(-1.7)}$ &  16.0$_{(-3.2)}$ & 13.3$_{(-5.9)}$ & 13.8$_{(-5.4)}$& 16.3$_{(-2.9)}$ & 14.5$_{(-4.7)}$\\
\multicolumn{1}{l}{Electronics}      & 21.4   & 16.8$_{(-4.6)}$ & 14.3$_{(-7.1)}$ & 13.2$_{(-8.2)}$ & 11.5$_{(-9.9)}$ & 11.5$_{(-9.9)}$& 16.8$_{(-4.6)}$ & 13.8$_{(-7.6)}$\\
\multicolumn{1}{l}{DVD}              & 19.9   & 16.0$_{(-3.9)}$ & 16.5$_{(-3.4)}$ & 14.5$_{(-5.4)}$ & 13.5$_{(-6.4)}$ & 12.0$_{(-7.9)}$& 14.3$_{(-5.6)}$ & 14.0$_{(-5.9)}$\\
\multicolumn{1}{l}{Kitchen}          & 20.1   & 16.8$_{(-3.3)}$ & 14.0$_{(-6.1)}$ & 13.8$_{(-6.3)}$ & 12.3$_{(-7.8)}$ & 11.8$_{(-8.3)}$& 15.0$_{(-5.1)}$ & 12.8$_{(-7.3)}$\\
\multicolumn{1}{l}{Apparel}          & 15.7   & 16.3$_{(+0.6)}$ & 15.5$_{(-0.2)}$ & 13.0$_{(-2.7)}$ & 13.0$_{(-2.7)}$ & 12.5$_{(-3.2)}$& 13.8$_{(-1.9)}$ & 13.5$_{(-2.2)}$\\
\multicolumn{1}{l}{Camera}           & 14.6   & 14.0$_{(-0.6)}$ & 13.5$_{(-1.1)}$ & 10.8$_{(-3.8)}$ & 10.5$_{(-4.1)}$ & 11.0$_{(-3.6)}$& 10.3$_{(-4.3)}$ & 11.0$_{(-3.6)}$\\
\multicolumn{1}{l}{Health}           & 17.8   & 12.8$_{(-5.0)}$ & 12.0$_{(-5.8)}$ & 11.8$_{(-6.0)}$ & 10.5$_{(-7.3)}$ & 10.5$_{(-7.3)}$& 13.5$_{(-4.3)}$ & 11.0$_{(-6.8)}$\\
\multicolumn{1}{l}{Music}            & 23.0   & 16.3$_{(-6.7)}$ & 18.8$_{(-4.2)}$ & 17.5$_{(-5.5)}$ & 14.8$_{(-8.2)}$ & 14.3$_{(-8.7)}$& 18.3$_{(-4.7)}$ & 14.8$_{(-8.2)}$\\
\multicolumn{1}{l}{Toys}             & 16.3   & 10.8$_{(-5.5)}$ & 15.5$_{(-0.8)}$ & 12.0$_{(-4.3)}$ & 11.0$_{(-5.3)}$ & 10.8$_{(-5.5)}$& 11.8$_{(-4.5)}$ & 10.8$_{(-5.5)}$\\
\multicolumn{1}{l}{Video}            & 17.0   & 18.5$_{(+1.5)}$ & 16.3$_{(-0.7)}$ & 15.5$_{(-1.5)}$ & 13.0$_{(-4.0)}$ & 14.0$_{(-3.0)}$& 14.8$_{(-2.2)}$ & 13.5$_{(-3.5)}$\\
\multicolumn{1}{l}{Baby}             & 15.9   & 12.3$_{(-3.6)}$ & 12.0$_{(-3.9)}$ & 11.8$_{(-4.1)}$ & 10.8$_{(-5.1)}$ & 11.3$_{(-4.6)}$& 12.0$_{(-3.9)}$ & 11.5$_{(-4.4)}$\\
\multicolumn{1}{l}{Magazines}        & 10.5   & 12.3$_{(+1.8)}$ &  7.5$_{(-3.0)}$ & 7.8$_{(-2.7)}$ &  8.0$_{(-2.5)}$ & 7.8$_{(-2.7)}$& 9.5$_{(-1.0)}$    &  8.8$_{(-1.7)}$\\
\multicolumn{1}{l}{Software}         & 14.7   & 13.5$_{(-1.2)}$ & 13.8$_{(-0.9)}$ & 12.8$_{(-1.9)}$ & 10.3$_{(-4.4)}$ & 12.8$_{(-1.9)}$& 11.8$_{(-2.9)}$ & 11.0$_{(-3.7)}$\\
\multicolumn{1}{l}{Sports}           & 17.3   & 16.0$_{(-1.3)}$ & 14.5$_{(-2.8)}$ & 14.3$_{(-3.0)}$ & 12.3$_{(-5.0)}$ & 13.3$_{(-4.0)}$& 13.5$_{(-3.8)}$ & 12.8$_{(-4.5)}$\\
\multicolumn{1}{l}{IMDB}             & 17.3   & 13.8$_{(-3.5)}$ & 17.5$_{(+0.2)}$ & 14.5$_{(-2.8)}$ & 13.0$_{(-4.3)}$ & 13.5$_{(-3.8)}$& 13.3$_{(-4.0)}$ & 13.3$_{(-4.0)}$\\
\multicolumn{1}{l}{MR}               & 26.9   & 25.5$_{(-1.4)}$ & 25.3$_{(-1.6)}$ & 23.3$_{(-3.6)}$ & 21.5$_{(-5.4)}$ & 22.0$_{(-4.9)}$& 23.5$_{(-3.4)}$ & 22.8$_{(-4.1)}$\\
\midrule
\rowcolor[gray]{.7}
\multicolumn{1}{l}{AVG}              & 18.0   & 15.5$_{(-2.5)}$ & 15.3$_{(-2.7)}$ & 13.9$_{(-4.1)}$  & \textbf{12.5$_{(-5.5)}$} & {12.7$_{(-5.3)}$} & 14.3$_{(-3.7)}$ & 13.1$_{(-4.9)}$\\
\bottomrule
\end{tabular}
\caption{
Text classification error rates of our models on 16 datasets against typical baselines. The smaller values indicate better performances.
The column of \textbf{Single Task} (Avg.) gives the average error rates of vanilla LSTM, bidirectional LSTM, and stacked LSTM while the column of \textbf{Multiple Tasks} shows the results achieved by corresponding multi-task models. The \textbf{Transfer} column lists the results of different models on transfer learning.
``*'' indicates our proposed models.
The numbers in brackets represent the improvements relative to the average performance (Avg.) of three single task baselines.
}\label{tab:mtl-16task}
\end{table*}

We choose several relevant and representative models as baselines.
\begin{itemize*}
    \item MT-CNN: This model is proposed by \citet{collobert2008unified} with a convolutional layer, in which lookup-tables are shared partially while other layers are task-specific.
    \item FS-MTL: Fully shared multi-task learning framework. Different tasks fully share a neural layer (LSTM).
    \item SP-MTL: Shared-private multi-task learning framework with adversarial learning \citep{liu2017adversarial}. Different tasks not only have common layers to share information, but have their own private layers.
\end{itemize*}

\textbf{Results on Multi-task Learning:}
The experimental results show that our proposed models outperform all single-task baselines by a large margin, and here we show the averaged error due to the following reasons: 1) it is easier to show the performance gain of multi-task learning models over single task models. 2) BiLSTM and stacked LSTM are also the necessary baselines for SG-MTL, since the combination of shared and private layers in SG-MTL is similar to two-layer LSTM.

Table~\ref{tab:mtl-16task} shows the overall results on the 16 different tasks under three settings: single task, multiple task, and transfer learning. Generally, we can see that almost all tasks benefit from multi-task learning, which boosts the performance by a large margin.
Specifically, CG-MTL achieves the best performance, surpassing SP-MTL by  1.4$\%$, which suggests that explicit communication makes it easier to shared information.
Although the improvement of SG-MTL is not as large as CG-MTL, SG-MTL is efficient to train.
Additionally, the comparison between  SG-MTL and SP-MTL shows the effectiveness of selectively sharing schema.
Moreover, we may further improve our models by incorporating the adversarial training mechanism introduced in SP-MTL, as it is an orthogonal innovation to our methods.

\textbf{Evaluation on Transfer Learning:}
We next present the potential of our methods on transfer learning, as we expect that the shared knowledge learned by our model architectures can be useful for new tasks.
In particular, the virtual node in the SG-MTL model can condense shared information into a common space after multi-task learning, which allows us to transfer this knowledge to new tasks.
In order to test the transferability of the shared knowledge learned by SG-MTL, we design an experiment following the supervised pre-training paradigm.
Specifically, we adopt a 16-fold ``leave-one-task-out'' paradigm; we take turns choosing $15$ tasks to train our model via multi-task learning, then the learned shared layer is transferred to a second network that is used to test on the remaining \emph{target task} $k$. The parameters of the transferred layer are kept frozen, and the remaining parameters of the new network are randomly initialized.

Table~\ref{tab:mtl-16task} shows these results in the ``\texttt{Transfer}'' column, in which the task in each row is regarded as the target task.
We observe that our model achieves a $4.9\%$ average improvement in terms of the error rate over the single tasking setting (13.1 vs. 18.0), surpassing SP-MTL by $1.2\%$ in average (13.1 vs. 14.3). This improvement suggests that our retrieval method with the selective mechanism (the attention layer in eq. \ref{eq:shared_lstm}) is more efficient in finding the relevant information from the shared space compared to SP-MTL,  
which reads the shared information without any selective mechanism and ignores the relationship between tasks.

\begin{figure*}[!t]
\setlength{\belowcaptionskip}{-0.1cm}
\centering
 \includegraphics[width=0.9\linewidth]{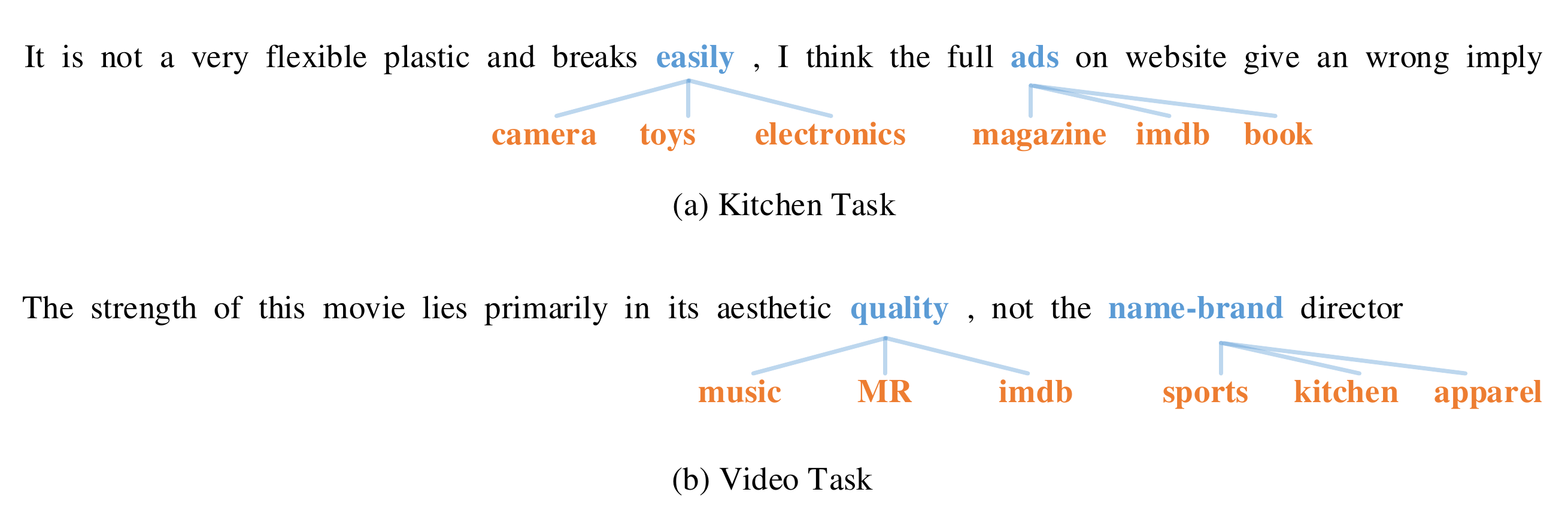}
 \caption{
  Illustrations of the most relevant tasks (top 3) for each word in ``\texttt{Kitchen}'' and ``\texttt{Video}'' tasks.
  Here we choose some typical words to visualize in blue. And the orange words represent the relevant tasks.
 }\label{fig:dc-exp}
\end{figure*}

\begin{table*}[!t]
\small

\centering
\begin{tabular}{llllll}
\toprule
  & \multicolumn{4}{c}{\textbf{Interpretable Sub-Structures}} & \textbf{Explanations}\\
\midrule
\textbf{Short-term}
& \raisebox{-.5\height}{\includegraphics[width=0.10\textwidth]{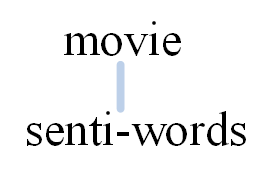}}
& \raisebox{-.5\height}{\includegraphics[width=0.14\textwidth]{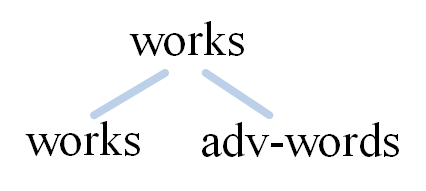}}
& \raisebox{-.5\height}{\includegraphics[width=0.09\textwidth]{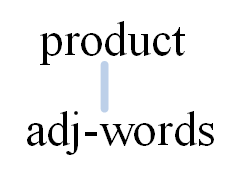}}
& \raisebox{-.5\height}{\includegraphics[width=0.12\textwidth]{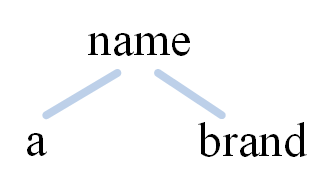}}
& \tabincell{l}{Interpretable phrases to be \\shared across tasks}
\\
\midrule
\textbf{Long-term}
& \raisebox{-.5\height}{\includegraphics[width=0.12\textwidth]{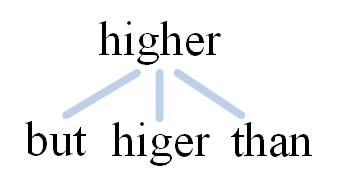}}
& \raisebox{-.5\height}{\includegraphics[width=0.13\textwidth]{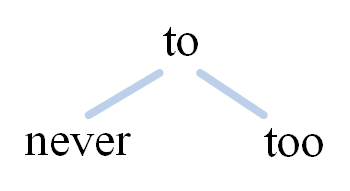}}
& \raisebox{-.5\height}{\includegraphics[width=0.12\textwidth]{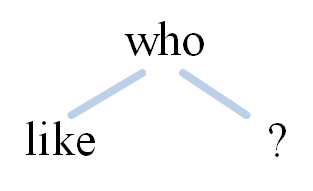}}
& \raisebox{-.5\height}{\includegraphics[width=0.12\textwidth]{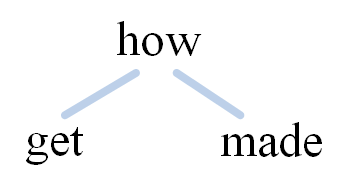}}
& \tabincell{l}{Interpretable sentence patterns, \\they usually determine the \\sentence meanings.}
\\
\bottomrule
\end{tabular}
\caption{Multiple interpretable sub-structures learned by shared layers.
``\texttt{senti-words}'' refers to ``\texttt{boring, interesting, amazing}'' etc.
``\texttt{adv-words}'' refers to ``\texttt{easily, fine, great}'' etc.
``\texttt{adj-words}'' refers to ``\texttt{stable, great, fantastic}'' etc.
These structures show that short-term and long-term dependencies between different words can be learned, which usually control the sentiment of corresponding sentences.
} \label{tab:id-exp1}
\end{table*}

\subsection{Sequence Tagging}
In this section, we present the results of our models on the second task of sequence tagging. We conducted experiments by following the same settings as \citet{yang2016multi}.
We use the following benchmark datasets in our experiments: Penn Treebank (PTB) POS tagging, CoNLL 2000 chunking, CoNLL 2003 English NER. The statistics of the datasets are described in Table~\ref{tab:st}.

\begin{table}[!t]\small
\center
\tabcolsep0.06in
\begin{tabular}{l*{6}{l}}
\toprule
\textbf{Dataset} &       \textbf{Task} &	 	 \textbf{Training } &	 \textbf{Dev. } &	 \textbf{Test } \\
\midrule

CoNLL 2000 &\multirow{1}{*}{Chunking}
&  211,727 &  - &  47,377    \\
\midrule
CoNLL 2003 &\multirow{1}{*}{NER}
&  204,567 &  51,578 &  46,666    \\
\midrule
PTB &  \multirow{1}{*}{POS}
&  912,344 &  131,768 &  129,654     \\

\bottomrule
\end{tabular}
\caption{The sizes of the sequence labelling datasets in our experiments, in terms of the number of tokens.}\label{tab:st}
\end{table}

\begin{table}[!t]
\small
\centering
\begin{tabular}{lccc}
\toprule
\textbf{Model} & CoNLL2000 & CoNLL2003 & PTB \\
\midrule
LSTM + CRF    & 94.46  & 90.10 & {97.55} \\
MT-CNN          & 94.32  & 89.59 & 97.29 \\
FS-MTL          & 95.41  & 90.94 & 97.55 \\
SP-MTL$^{\ast}$        & 95.27  & 90.90 & 97.49\\
\midrule
CG-MTL        & 95.49  & 91.25 & 97.61 \\
SG-MTL        & \textbf{95.61}  & \textbf{91.47} & \textbf{97.69} \\
\bottomrule
\end{tabular}
\caption{Performance of different models on Chunking, NER, and POS respectively.
All the results without marks are reported in the corresponding paper.
LSTM+ CRF: Proposed by \citep{huang2015bidirectional}.
MT-CNN: Proposed by \citep{collobert2008unified}.
FS-MTL: Proposed by  \citep{yang2016multi}.
} \label{tab:exp-st}
\end{table}

\textbf{Results and Analysis:}
Table~\ref{tab:exp-st} shows the performance of the models on the sequence tagging tasks. CG-MTL and SG-MTL significantly outperform the three strong multi-task learning baselines,  Specifically, SG-MTL achieves a performance gain of $0.53\%$ in terms of F1 score over the best competitor FS-MTL on the CoNLL2003 dataset,  indicating that our models are able to make use of the shared information by modelling the relationship between different tasks. Our models also achieve slightly better F1 scores on the CoNLL2000 and PTB datasets when compared to the best baseline model FS-MTL.

\section{Discussion and Qualitative Analysis}
In order to obtain more insights and detailed interpretability of how messages are passed between tasks in our proposed models, we design a series of experiments targeting the following aspects:

\begin{enumerate}
    \item Can the relationship between different tasks be learned by CG-MTL?
    \item Are there interpretable structures that the shared layer in SG-MTL can learn?  Are these shared patterns similar to linguistic structures, and can they be transferred for other tasks? 
\end{enumerate}

\paragraph{Explicit Relationship Learning}

To answer the first question, we visualize the weight $\alpha_{t}$ of CG-MTL in equation \ref{eq:dc_mtl}. As each task can receive messages from any other task in CG-MTL, $\alpha_{t}$  directly indicates the relevance of other tasks to the current task at time step $t$.
As shown in Figure~\ref{fig:dc-exp}, we analyze the relationships learned by our models on randomly sampled sentences from different tasks.
We find that the relationship between tasks cannot be modelled by a static score. Rather, it depends on the specific sample and context. Consider the example sentence in Figure~\ref{fig:dc-exp}-(a), drawn from the \textsc{Kitchen} task. Here, the words ``\texttt{easily}'' and ``\texttt{ads}''  are influenced by different sets of external tasks, in which those words express sentiment. For example, in the \textsc{Camera} and \textsc{Toys} tasks, ``\texttt{breaks easily}'' is usually used to express negative sentiment, while the word ``\texttt{ads}'' often appears in the \textsc{Magazine} task to express negative sentiment. Figure \ref{fig:dc-exp}-(b) shows a similar case on ``\texttt{quality}'' and  ``\texttt{name-brand}''.

\begin{figure}[!t]
\setlength{\belowcaptionskip}{-0.1cm}
\centering
 \includegraphics[width=0.88\linewidth]{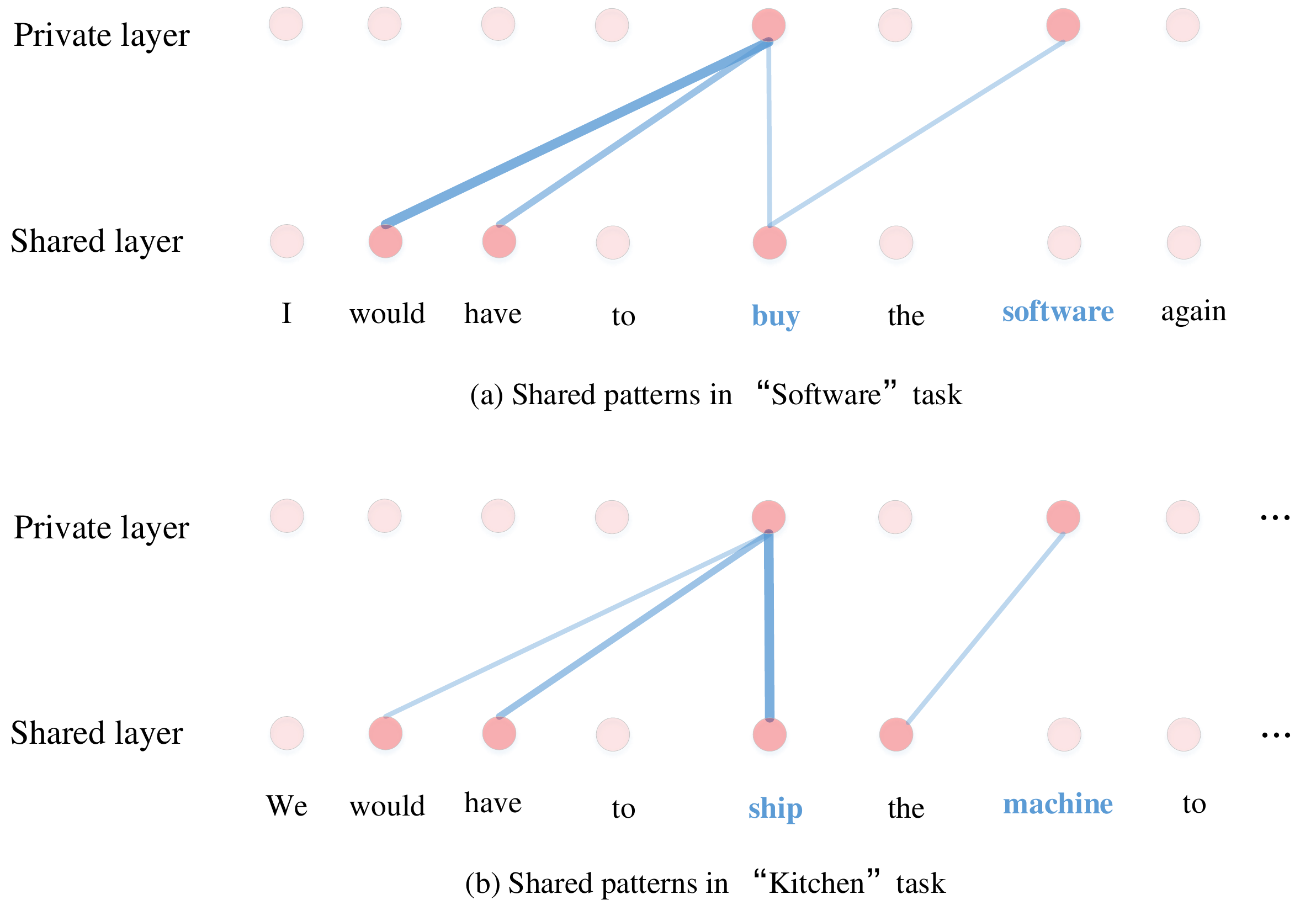}
 \caption{
  Illustrations of the learned patterns captured by the shared layer under different tasks. The boldness of the line indicates the strength $\beta$ of the relationship.
  The first sentence comes from the development set of task \textsc{Software} while the second one belongs to the training set of task \textsc{Kitchen}.
  We choose three typical words ``\texttt{buy}'', ``\texttt{software}'' and ``\texttt{machine}'' to visualize in these sampled sentences.
 }\label{fig:id-exp1}
\end{figure}

\paragraph{Interpretable Structure Learning}
To answer the second question, we visualize $\beta$ in equation \ref{eq:shared_lstm} inside the SG-MTL model. As different tasks can read information from shared layers in SG-MTL, visualizing $\beta$  allows us to analyze what kinds of sentence structures are shared.
Specifically, each word $w_{t}^{(k)}$ can receive shared messages: $w_{1}^{(s)}$ $..$. $w_{T}^{(s)}$  and the amount of messages is controlled by the scores $\beta$. To illustrate the interpretable structures learned by the shared layer in SG-MTL, we randomly sample several examples from different tasks and visualize their shared structures. Three random sampled cases are described as in Figure \ref{fig:id-exp1}.

From the experiments we conducted in visulizing $\beta$ in SG-MTL, we observed the following:
\begin{itemize}
    \item The proposed model can not only utilize the shared information across different tasks, but can tell us what kinds of features are shared.
    As shown in Table~\ref{tab:id-exp1}, the short-term and long-term dependencies between different words can be captured.
    For example, the word ``\texttt{movie}'' is prone to connecting to emotional words, such as  ``\texttt{boring, amazing, exciting}'' while ``\texttt{products}'' is more likely to make friends with ``\texttt{stable, great, fantastic}''.
    \item Comparing Figure~\ref{fig:id-exp1}-(b) and (c), we can see how task \textsc{Software}  borrows useful information from task \textsc{Kitchen}.
    Concretely, the sentence ``\texttt{I \textbf{would have} to buy the software again}'' in the task ``\texttt{Software}'' has negative emotion. In this sentence, the key pattern is ``\texttt{would have}'', which does not appear too much in the training set of \textsc{Software}. Fortunately, the training samples in the task \textsc{Kitchen} provide more hints about this pattern.

    \item As shown in Figure~\ref{fig:id-exp1}-(a) and (b), the shared layer has learned an informative sentence pattern ``\texttt{would have to ...}'' from the training set of task \textsc{Kitchen}. This pattern is useful for the sentiment prediction of another task \textsc{Software}, which suggests that we can analyze the sharabla patterns in an interpretable way for SG-MTL model.
\end{itemize}

\section{Related Work}
Neural network-based multi-task frameworks have achieved success on many NLP tasks, such as POS tagging \citep{yang2016multi,sogaard2016deep}, parsing \citep{peng2017deep,guo2016exploiting}, machine translation \citep{dong2015multi,luong2015multi,firat2016multi}, and text classification \citep{liu2016recurrent,liu2017adversarial}. However, previous work does not focus on explicitly modelling the relationships between different tasks. These models are often trained with an opaque neural component, which makes it hard to understand what kind of knowledge is shared. 
By contrast, in this paper, we propose to explicitly learn the communication between different tasks, and learn some interpretable shared structures.

Before the bloom of neural-based models, \textbf{non-neural multi-task learning} methods have also been proposed to model the relationships between tasks.
For example, \citet{bakker2003task} learn to cluster tasks by using Bayesian approaches. \citet{kim2010tree} utilizes a given tree structure to design a regularizer, while \citet{chen2010graph} learns a structured multi-task problem over a given graph.
These models adopt complex learning strategies and introduce a priori information between different tasks, which are usually not suitable for sequence modelling.
In this paper, we provide a new perspective on how to model the relationships using distributed graph models and \textit{message passing}, which can be learned dynamically rather than following a pre-defined structure.

The technique of message passing is used ubiquitously in computer software \citep{berendsen1995gromacs} and programming languages \citep{serlet1996method}.
Recently, there has also been growing interest in developing graph neural networks \citep{kipf2016semi} or neural message passing algorithms \citep{gilmer2017neural} for learning representations of irregular graph-structured data.
In this paper, we formulate  multi-task learning as a communication problem over graph structures, allowing different tasks to communicate via message passing.

More recently, \cite{liu2018meta} propose to learn multi-task communication by explicitly passing gradients. Both our work try to incorporate inductive bias to multi-task learning. However, the difference is that we focus on the structural bias while \cite{liu2018meta} introduced an additional loss function.

\section{Conclusion and Outlook}
We have explored the problem of learning the relationships between multiple tasks, formulating the problem as message passing over a graph neural network.
Our proposed methods explicitly model the relationships between different tasks and achieve improved performance in several multi-task and transfer learning settings. We also show that we can extract interpretable shared patterns from the outputs of our models. From our experiments, we believe that learning interpretable shared structures is a promising direction, which is also very useful for knowledge transfer.

\section*{Acknowledgments}
The authors wish to thank the anonymous reviewers for their helpful comments. This work was partially funded by National Natural Science Foundation of China (No. 61751201, 61672162), STCSM (No.16JC1420401, No.17JC1404100), and Natural Sciences and Engineering Research Council of
Canada (NSERC).

\bibliographystyle{aaai}
\bibliography{./nlp1,./ours1,./nlp,./ours2}


\end{document}